\documentclass[conference]{IEEEtran}
\usepackage{cite}
\usepackage{amsmath,amssymb,amsfonts}
\usepackage{algorithm}
\usepackage{algpseudocode}
\usepackage{graphicx}
\usepackage{textcomp}
\usepackage{xcolor}
\usepackage{booktabs} 
\usepackage{caption}
\def\BibTeX{{\rm B\kern-.05em{\sc i\kern-.025em b}\kern-.08em
    T\kern-.1667em\lower.7ex\hbox{E}\kern-.125emX}}
\begin{document}

\title{AW-GATCN: Adaptive Weighted Graph Attention Convolutional Network for Event Camera Data Joint Denoising and Object Recognition\\
}

\author{\IEEEauthorblockN{1\textsuperscript{st} Haiyu Li}
\IEEEauthorblockA{\textit{School of Electronic and Electrical Engineering} \\
\textit{The University of Sheffield}\\
Sheffield, United Kingdom \\
hli108@sheffield.ac.uk}
\and
\IEEEauthorblockN{2\textsuperscript{nd} Charith Abhayaratne}
\IEEEauthorblockA{\textit{School of Electronic and Electrical Engineering} \\
\textit{Centre for Machine Intelligence} \\
\textit{The University of Sheffield}\\
Sheffield, United Kingdom  \\
c.abhayaratne@sheffield.ac.uk}
}

\maketitle

\begin{abstract}
Event cameras, which capture brightness changes with high temporal resolution, inherently generate a significant amount of redundant and noisy data beyond essential object structures. The primary challenge in event-based object recognition lies in effectively removing this noise without losing critical spatial-temporal information. To address this, we propose an Adaptive Graph-based Noisy Data Removal framework for Event-based Object Recognition. Specifically, our approach integrates adaptive event segmentation based on normalized density analysis, a multifactorial edge-weighting mechanism, and adaptive graph-based denoising strategies. These innovations significantly enhance the integration of spatiotemporal information, effectively filtering noise while preserving critical structural features for robust recognition. Experimental evaluations on four challenging datasets demonstrate that our method achieves superior recognition accuracies of 83.77\%, 76.79\%, 99.30\%, and 96.89\%, surpassing existing graph-based methods by up to 8.79\%, and improving noise reduction performance by up to 19.57\%, with an additional accuracy gain of 6.26\% compared to traditional Euclidean-based techniques.
\end{abstract}
\footnotetext{This is the preprint version of a paper accepted at IJCNN 2025. The final version will appear in the IJCNN 2025 proceedings.}


\section{Introduction}
The advent of artificial intelligence has propelled computer vision to the forefront of real-world applications such as autonomous driving, drone navigation, and surveillance, where swift and accurate object recognition is paramount. Traditional video cameras, limited by low frame rates and excessive data redundancy, fall short in these dynamic settings. Although high-speed cameras capture over 1,000 frames per second, their high cost limits practicality. In contrast, event cameras\cite{b1}, which only record changes in scene brightness, significantly minimize data redundancy and are unaffected by motion blur. They offer microsecond-level temporal resolution and low latency, making them ideal for environments that demand rapid and reliable data processing.

Unlike conventional cameras that output continuous two-dimensional images, event cameras are triggered by significant changes in pixel brightness, efficiently eliminating most of irrelevant background information. However, the asynchronous and sparse data they generate pose significant challenges for traditional frame-based processing techniques\cite{b2}. Researchers typically convert event streams into 2D frames or 3D voxel grids\cite{b3,b4,b5}, a process that compromises the data's inherent sparsity and temporal resolution, leading to potential information loss. The absence of a standardized conversion method further complicates data processing, as application-specific needs require customized approaches, yielding inconsistent results across different scenarios.

To fully leverage the unique characteristics of event cameras—namely, the sparsity and asynchrony of event data—researchers have explored innovative processing methodologies such as temporal surface-based methods\cite{b6,b7} and spiking neural networks (SNNs)\cite{b8,b9}. These approaches, which process data on an event-by-event basis, are designed to maintain low latency. However, their efficacy in complex tasks can be limited due to the sensitivity to parameter settings and the intricacies of their training processes. In response to these challenges, recent advancements have introduced a more efficient strategy involving compact graph representations\cite{b10,b12,b13}. These methods model event sequences as graphs within a cloud of events using graph convolutional networks (GCNs), which have achieved state-of-the-art performance. Despite their successes, these graph-based approaches primarily rely on a simplistic radius-based noise management strategy—connecting nodes only if they are within a predetermined Euclidean distance. This technique often proves inadequate for effectively handling noise and lacks the flexibility needed for adapting to dynamically changing environments.

Based on these observations, we propose an adaptive graph formulation-based noise reduction algorithm integrated with a graph convolutional neural network (GCN) that incorporates attention mechanisms, enabling efficient and accurate processing of event data. Traditional radius-based methods rely on fixed Euclidean distances, limiting adaptability and robustness while overlooking other informative graph features. Our approach overcomes these limitations by incorporating multilevel weights, dynamically adjusting weight thresholds and leveraging a graph attention mechanism for enhanced feature aggregation and classification.

The main contributions of this paper are as follows:

\begin{itemize}
   \item \textbf{Multi-Factor Edge Weighting:} A robust edge weighting mechanism that incorporates Euclidean distance, velocity, angular difference, and polarity consistency, ensuring accurate modeling of event point relationships, even in noisy environments.
   
   \item \textbf{Adaptive graph formulation-based Noise Reduction:} A dynamic noise reduction strategy that adapts the formulation of the underlying graph of event based on the variance in node distribution, effectively filtering out noisy events by preserving the key event data in sparse areas and removing excess in dense regions.
   
   \item \textbf{Graph Attention Convolutional Network:} A GCN guided by multi-factor edge weighting, selectively emphasizing relevant neighboring features and achieving enhanced data representation and object recognition accuracy.

\end{itemize}

\section{Related Work}
Current methods for event data processing can be broadly categorized into frame-based conversion methods, graph-based methods, and deep learning methods. Frame-based methods, such as those proposed by\cite{b3,b4}, convert event data into pseudo-images, but this often leads to a loss of temporal resolution. Graph-based methods\cite{b15,b16} maintain the sparsity of event data but face challenges in processing efficiency. Deep learning methods, particularly those utilizing Graph Attention Neural Networks\cite{b17}, are adept at handling complex graph-structured data but often struggle with noisy data\cite{b18}. 

One approach for event cameras is to use Spiking Neural Networks (SNN)\cite{b19}, again a biologically inspired design.SNNs exploit the sparsity and asynchronous nature of event data, but due to their non-microscopic nature, training such networks is very difficult. To improve the temporal resolution, Zhu et al\cite{b21,b22} suggested discretizing the temporal dimension into consecutive time segments and accumulating the events into a voxel grid by linear weighted accumulation similar to bilinear interpolation. Messikommer et al\cite{b23} further exploited spatial and temporal sparsity by employing sparse convolution\cite{b24} and developing recursive convolution formulations. However, they still operate on sparse volumes and 3D convolution is computationally expensive for dealing with large event clouds.

Recent studies, e.g., further use a framework similar to PointNet\cite{b25,b26}, which utilizes a multilayer perceptron (MLP) to learn the features of each point separately and then outputs object-level responses (e.g., categorical labels) via a global max operation. For event processing, Sekikawa et al\cite{b27} developed for the first time a recursive architecture called EventNet. Specifically, it recursively represents the dependencies of causal events on outputs through a new temporal encoding and aggregation scheme and pre-computes the features of nodes that correspond to particular spatial coordinates and polarities.

\begin{figure*}[!t]
    \centering
    \begin{minipage}{\textwidth}
        \centering
        \includegraphics[width=0.85\linewidth]{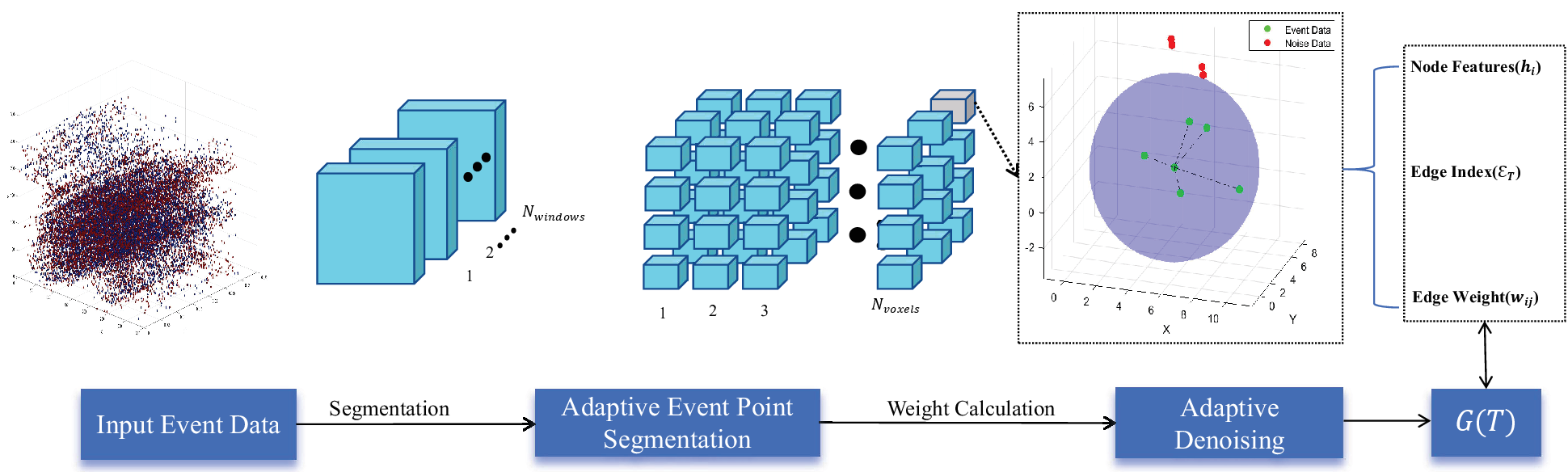} 
        \vspace{4mm} 
        \includegraphics[width=0.85\linewidth]{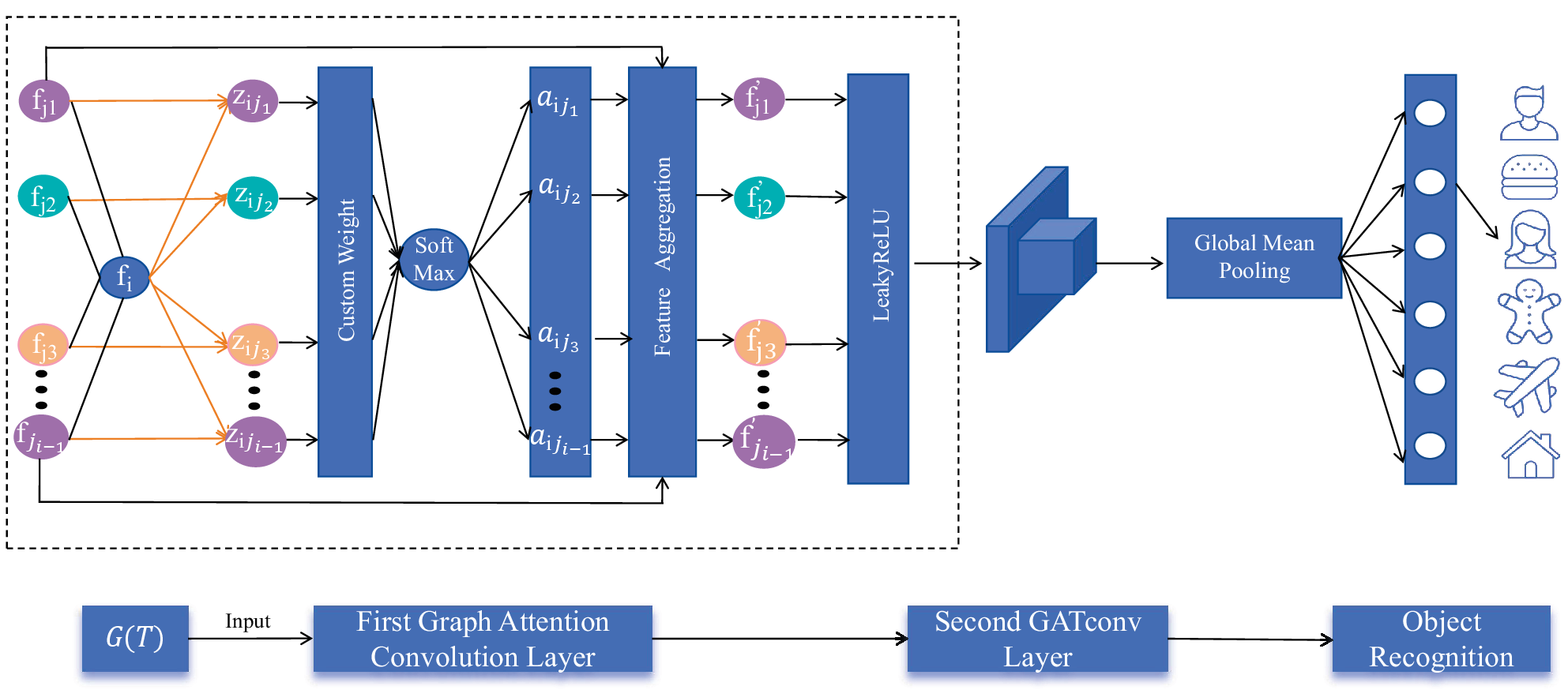} 
    \end{minipage}
    \caption{The adaptive edge weight \( w_{ij} \), designed to capture event point relevance, facilitates noise filtering and attention-based feature aggregation, enhancing robust recognition by emphasizing the most informative node connections.}
    \label{fig:framework}
    \vspace{-2mm} 
\end{figure*}

\section{Methodology}

Processing event videos from event cameras requires managing large volumes of noisy event points\cite{b28}, making direct processing computationally demanding. To address this, we propose an adaptive segmentation algorithm that first divides the input data into balanced windows based on normalized density and then subdivides each window into voxels using the square root law to balance temporal and spatial dimensions. For noise reduction, inspired by\cite{b29},  
we employ an adaptive algorithm that dynamically adjusts the weighting radius based on multiple event point features, filtering out noise. These weights are then integrated with a graph attention mechanism to selectively focus on relevant neighboring features, improving object recognition performance. An overview of the framework is shown in Figure~\ref{fig:framework}.

\subsection{Adaptive Event Point Segmentation}

Event data generated by event cameras is represented as a point cloud, denoted as \(\mathcal{E} = \{(x_k, y_k, t_k, p_k)\}_{k=1}^{N_{\text{points}}}\), where each point \(k\) includes spatial coordinates \((x_k, y_k)\), a timestamp \(t_k\), and a polarity \(p_k\). Here, \(N_{\text{points}}\) represents the total number of event points. Algorithm 1 outlines the preprocessing and segmentation procedures, which are conducted as follows:

\begin{algorithm}[t]
\caption{Adaptive Event Point Segmentation}
\label{alg:adaptive_event_point_segmentation}
\begin{algorithmic}[1]
\State $x_{\text{range}} \gets x_{\max} - x_{\min}$
\State $y_{\text{range}} \gets y_{\max} - y_{\min}$
\State $t_{\text{range}} \gets t_{\max} - t_{\min}$
\State $\rho_{\text{norm}} \gets \frac{N_{\text{points}}}{x_{\text{range}} \times y_{\text{range}} \times t_{\text{range}}}$
\State $N_{\text{window}} \gets \max(N_{\text{min}}, \rho_{\text{norm}} \times C_{\text{scale}})$
\State $N_{\text{windows}} \gets \left\lceil \frac{N_{\text{points}}}{N_{\text{window}}} \right\rceil$
\For{$w \gets 1$ to $N_{\text{windows}}$}
    \State Sort data in window $w$ by $t$
    \State $x_{\text{span}} \gets x_{\max} - x_{\min}$
    \State $y_{\text{span}} \gets y_{\max} - y_{\min}$
    \State $t_{\text{span}} \gets t_{\max} - t_{\min}$
    \State $N_{\text{voxels}} \gets \sqrt{x_{\text{span}} \times y_{\text{span}} \times t_{\text{span}}}$
    \State $N_{\text{voxels}} \gets \max(N_{\text{min\_vox}}, \min(N_{\text{max\_vox}}, N_{\text{voxels}}))$
    \State Divide window $w$ into $N_{\text{voxels}}$ based on time intervals
    \For{$m \gets 1$ to $N_{\text{voxels}}$}
        \State Extract event points in voxel $m$
    \EndFor
\EndFor
\end{algorithmic}
\end{algorithm}

\subsubsection{Normalized Density}

To effectively segment event-based data, we commence by calculating the normalized density of data points. This calculation begins with determining the spatial range, capturing the data's extent in the \(x\) and \(y\) dimensions. By evaluating the distribution of data points along these axes, we establish the spatial boundaries essential for understanding the overall spatial distribution.

We then assess the temporal range of the data, reflecting the duration over which the events are recorded. This temporal assessment aids in comprehending the temporal dynamics of the event stream, enhancing our understanding of its temporal characteristics.

With both spatial and temporal ranges defined, the normalized density is computed as the number of data points per unit volume, encompassing both spatial and temporal dimensions. The formula for normalized density is given by:
\begin{equation}
\rho_{\text{nor}} = \frac{N_{\text{points}}}{(x_{\text{max}} - x_{\text{min}}) \cdot (y_{\text{max}} - y_{\text{min}}) \cdot (t_{\text{max}} - t_{\text{min}})},
\end{equation}
where \(N_{\text{points}}\) denotes the total number of data points, \((x_{\text{max}}, x_{\text{min}}, y_{\text{max}}, y_{\text{min}})\) represent the spatial boundaries, and \(t_{\text{max}}, t_{\text{min}}\) are the temporal boundaries.

\subsubsection{Determination of Adaptive Window Size}

To ensure an even distribution of data points across windows, we adaptively adjust the window size based on the normalized density. The number of points per window is determined by the following heuristic:
\begin{align}
n_{\text{window}} = \max\left( N_{\text{min}}, \rho_{\text{normalized}} \cdot C_{\text{scale}} \right),
\end{align}
where \( N_{\text{min}} \) is the minimum number of points per window, ensuring that windows are not overly sparse, and \( C_{\text{scale}} \) is a scaling factor to control the overall window size. This adjustment guarantees that each window contains a sufficient number of points for robust analysis, mitigating issues such as noise interference or sparse data regions.

After determining the optimal number of points per window, the total number of windows required to cover all data points is calculated as:
\begin{align}
N_{\text{windows}} = \left\lceil \frac{N_{\text{points}}}{n_{\text{window}}} \right\rceil,
\end{align}
where \( \lceil \cdot \rceil \) denotes the ceiling function, ensuring complete coverage of all data points across the windows.

This balanced distribution of data points within each window supports adaptive segmentation and facilitates reliable processing across varying data densities. The event point set \(\mathcal{E}\) is divided into multiple windows \(\{\mathcal{W}_w\}_{w=1}^{N_{\text{windows}}}\).

\subsubsection{Determination of Adaptive Voxel Count}

To further refine segmentation within each window, the voxel count is determined based on the \textbf{square root law}, which balances both spatial and temporal spans. The square root law is a heuristic often used in multidimensional systems to proportionally balance different dimensions by taking the square root of their product. In our approach, this allows us to determine an optimal voxel count that preserves data resolution while maintaining computational efficiency.

The number of voxels within each window is calculated as:
\begin{align}
N_{\text{voxels}} = \sqrt{(X) \cdot (Y) \cdot (T)},
\end{align}
where \( X = x_{\text{max}} - x_{\text{min}} \), \( Y = y_{\text{max}} - y_{\text{min}} \), and \( T = t_{\text{max}} - t_{\text{min}} \) represent the spatial and temporal ranges of the data within each window. This formula ensures that the voxelization process captures essential data characteristics. By applying the square root law, we achieve a segmentation that is both efficient and sufficiently detailed for subsequent analysis.

After dividing each window \(\mathcal{W}_w\) into multiple voxels \(\{V_m\}_{m=1}^{N_{\text{voxels}}}\), each voxel \(V_m\) contains a subset of event points characterized by their spatial, temporal, and polarity attributes. This segmentation enables each voxel \(V_m\) to encapsulate a localized subset of events, represented as follows:

\begin{align}
V_m = \{(x_k, y_k, t_k, p_k) \}.
\end{align}

Here, \(k\) represents the index of event points within voxel \(V_m\), and its range is determined by the number of points that fall within \(V_m\). Specifically, if \(V_m\) contains \(N_{\text{points}, m}\) event points, then \(k = 1, 2, \dots, N_{\text{points}, m}\).

\subsection{Adaptive Denoising}

AW-GATCN applies a weight-based noise reduction method that uses the variance of the normalized degree matrix to improve 3D structure recognition. This approach considers Euclidean distance, angular velocity difference, velocity magnitude difference, and polarity consistency between nodes. The optimal weight threshold is adaptively determined by maximizing the variance in node distribution across graphs within each voxel, unlike traditional methods with manually set thresholds. To maintain clarity and follow common graph processing conventions, we use \(k\) to denote event point indices during segmentation, switching to \(i\) and \(j\) for node indices in subsequent graph-based steps.

\subsubsection{Custom Weight Calculation}

For each pair of event points \(i\) and \(j\) within voxel \(V_m\), the edge weight \(w_{ij}\) is computed as:

\begin{align}
w_{ij} = \alpha \cdot D_{ij} + \beta \cdot \Delta v_{ij} + \gamma \cdot \theta_{ij} + \delta \cdot P_{ij},  \label{eq:weight_formula}
\end{align}

\begin{itemize}
    \item \( D_{ij} \): The Euclidean distance between nodes \(i\) and \(j\), representing the spatial distance.
    \item \( \Delta v_{ij} \): The magnitude difference between the velocity vectors of nodes \(i\) and \(j\).
    \item \( \theta_{ij} \): The angle difference between the velocity vectors or planar vectors of nodes \(i\) and \(j\).
    \item \( P_{ij} \): Polarity consistency, a binary indicator representing whether the polarities of nodes \(i\) and \(j\) are consistent (0 if consistent, 1 if inconsistent).
\end{itemize}

The velocity vector \(\mathbf{v}_i\) for each event point is computed based on the spatial and temporal differences between neighboring event points. For two event points \(e_i = (x_i, y_i, t_i)\) and \(e_j = (x_j, y_j, t_j)\), the velocity vector is defined as:

\begin{align}
\mathbf{v}_i = \left( \frac{x_j - x_i}{t_j - t_i}, \frac{y_j - y_i}{t_j - t_i} \right). \label{eq:velocity_vector}
\end{align}

This vector represents the "movement" of event point \(e_i\) through space and time. When \(t_i = t_j\), only the angular and magnitude differences in the 2D plane velocity vectors are calculated; if \(t_i \neq t_j\), both spatial and temporal velocity differences are computed. Since polarity does not influence velocity vector calculations, it is omitted here.

The angular difference between two velocity vectors describes their variation in motion direction and is given by:

\begin{align}
\cos \theta_{ij} = \frac{\mathbf{v}_i \cdot \mathbf{v}_j}{|\mathbf{v}_i||\mathbf{v}_j|},
\end{align}
where \(\mathbf{v}_i \cdot \mathbf{v}_j\) denotes the dot product, and \(|\mathbf{v}_i|\), \(|\mathbf{v}_j|\) are the magnitudes of the vectors. A smaller angle \(\theta_{ij}\) implies similar motion directions, while a larger angle indicates more pronounced directional differences.

In the weight calculation (see Equation~\eqref{eq:weight_formula}), the angular difference between velocity vectors plays a crucial role, as \(\theta_{ij}\) captures the similarity in motion directions. This term aids in identifying local motion patterns, such as sliding or rotating edges. By incorporating angular differences into the weight function, the method more accurately captures dynamic relationships between event points, thereby enhancing overall performance.

After conducting experiments, we finalized the weight parameters \(\alpha = 0.7\), \(\beta = 0.1\), \(\gamma = 0.1\), and \(\delta = 0.1\), as these values demonstrated effective performance in capturing dynamic relationships between event points (see Section~\ref{sec:Determination of Weight Parameters}).

\begin{figure}[t]
\centering
\includegraphics[width=\linewidth]{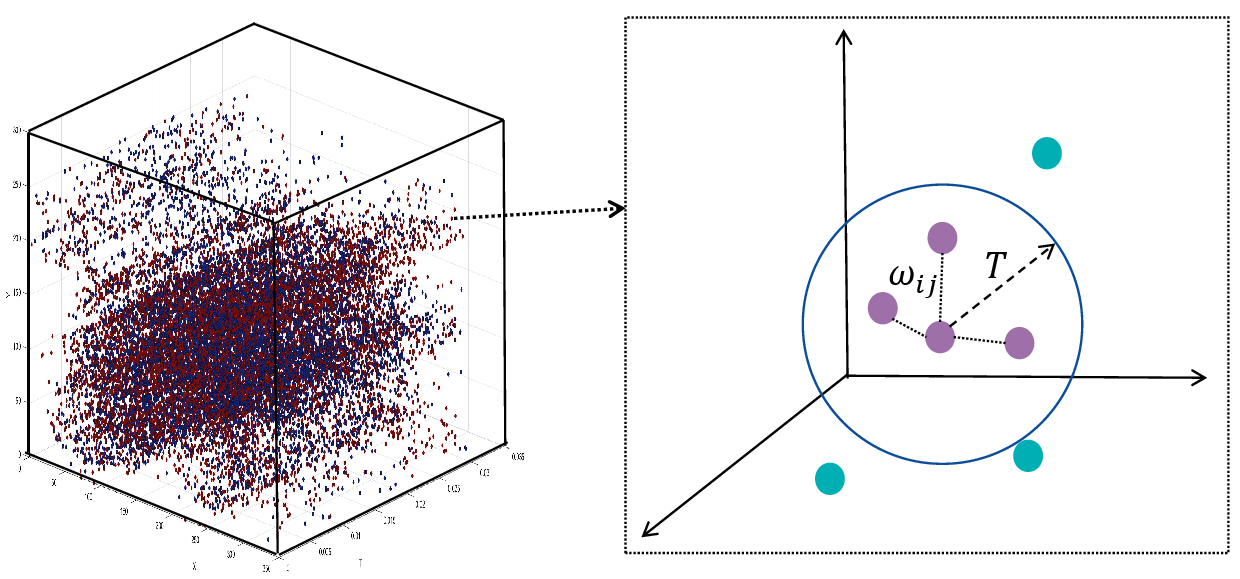}
\caption{The circular region represents the optimal threshold $T$ determined by maximizing variance. Purple nodes indicate high-correlation points with weights $w_{ij}$ less than $T$, while teal nodes are excluded.}
\label{fig:threshold_weight_relationship}
\end{figure}

\begin{algorithm}[ht]
\caption{Computing Optimal Threshold $T$}
\label{alg:optimal_threshold_awgatcn}
\begin{algorithmic}[1]
\For{$n = 0$ to $m - 1$} \Comment{$m$ is the number of voxels}
    \State $V_n \gets$ Voxel $n$ containing event points
    \State $t \gets$ Upper threshold for MST of $V_n$
    \For{$\delta = 0$ to $t$} \Comment{Threshold range from 0 to $t$}
        \State $\Phi_{\delta} \gets$ Node degree distribution at threshold $\delta$
        \State $\hat{\Phi}_{\delta} \gets \Phi_{\delta} / \max(\Phi_{\delta})$ \Comment{Normalize distribution}
        \State $\sigma^2_{\delta} \gets$ Variance of $\hat{\Phi}_{\delta}$
    \EndFor
    \State $T_n \gets \arg\max_{\delta} (\sigma^2_{\delta})$ \Comment{Maximize variance}
\EndFor
\State \textbf{return} $\{T_n\}_{n=0}^{m-1}$ \Comment{Set of optimal thresholds for all voxels}
\end{algorithmic}
\end{algorithm}

\subsubsection{Dynamic Threshold Adjustment Based on Normalized Degree Distribution}

\hspace{1em}Each voxel contains both event points and noise. To establish full connectivity for noise reduction, we first apply the Minimum Spanning Tree (MST) algorithm to determine the minimum weight that connects all event points within the voxel. This weight serves as the upper limit for the noise reduction threshold, ensuring that all points are fully connected. To effectively filter out noise while preserving significant connections, the threshold \(\delta\) is dynamically adjusted based on the normalized degree distribution of the graph. This process involves calculating the degree distribution for each voxel at various values of \(\delta\), where the distribution captures the number of connections within each voxel and reflects its relevance in the graph structure. This process is illustrated in Algorithm 2

The degree distribution is then normalized to a range of \([0, 1]\) to ensure consistency across different scales. This normalization is achieved by dividing the degree distribution \(\Phi_\delta\) by its maximum value:

\begin{align}
\Phi_\delta = \frac{\Phi_\delta}{\max(\Phi_\delta)}.
\end{align}

After normalization, the variance \(\sigma_\delta^2\) of the normalized degree distribution is computed for each threshold \(\delta\). This variance reflects the spread of the degree distribution, providing insight into the diversity of voxel connections within the graph.

The optimal threshold \(T\) is determined by selecting the value of \(\delta\) that maximizes the variance \(\sigma_\delta^2\), formulated as:

\begin{align}
T = \arg \max_\delta (\sigma_\delta^2).
\end{align}

Selecting the threshold that maximizes variance ensures that \(T\) preserves the most meaningful connections between nodes while effectively filtering out noise. 

\hspace{1em}The denoised graph \( G(T) = (\mathcal{V}, \mathcal{E}_T) \) is generated by retaining only edges with weights \( w_{ij} \leq T \). This filtering process removes irrelevant edges, enhancing the graph’s structure. The components of \( G(T) \) are:

\begin{itemize}
    \item \textbf{Nodes (Event Points):} Each node \( i \in \mathcal{V} \) represents an event point with feature vector:
    \begin{align}
    f_i = (x_i, y_i, t_i, p_i).
    \end{align}

    \item \textbf{Edges (Connections):} The edge set \( \mathcal{E}_T \) contains pairs \( (i, j) \) satisfying:
    \begin{align}
    \mathcal{E}_T = \{(i, j) \mid w_{ij} \leq T\}.
    \end{align}
    \item \textbf{Edge Weights:} Each edge weight \( w_{ij} \)(see Equation~\eqref{eq:weight_formula})
\end{itemize}

\subsection{AW-GATCN Network Architecture}

\subsubsection{Graph Convolutional Layer with Attention Mechanism}

\hspace{1em}In AW-GATCN, the graph convolutional layer employs an attention mechanism that dynamically adjusts edge weights, emphasizing the most relevant connections. The attention weight \(\alpha_{ij}\) for each edge between nodes \(i\) and \(j\) is computed based on the features of the target node \(f_i\) and its neighboring node \(f_j\), using a combination of learnable attention parameters and edge weights. In this framework, smaller edge weights indicate stronger correlations, prompting an inverse adjustment in the attention coefficient \(\alpha_{ij}\) based on edge weight, thereby enhancing the model's ability to capture meaningful relationships between nodes.

\begin{align}
\alpha_{ij} = \frac{\exp\left(\sigma\left(a^T [z_{ij}] \cdot \frac{1}{w_{ij}}\right)\right)}{\sum_{k \in \mathcal{N}(i)} \exp\left(\sigma\left(a^T [z_{ik}] \cdot \frac{1}{w_{ik}}\right)\right)},
\end{align}
where \(\sigma\) is the LeakyReLU activation function, \(a\) is a learnable parameter vector that captures the importance of neighboring nodes, and \(z_{ij} = Wf_i \parallel Wf_j\) represents the concatenation of the transformed feature vectors of nodes \(i\) and \(j\) via the weight matrix \(W\). The edge weight \(w_{ij}\), calculated through our adaptive weighting method, incorporates \(\frac{1}{w_{ij}}\) to emphasize edges with stronger correlations (i.e., smaller weights), thereby improving feature aggregation. In the denominator, \(\mathcal{N}(i)\) denotes the set of neighboring nodes of \(i\), and \(k\) iterates over each neighboring node in \(\mathcal{N}(i)\) for normalization.

The attention coefficients \(\alpha_{ij}\) are then used to aggregate features from neighboring nodes, improving each node’s representation by focusing on the connections with the highest correlation. The aggregated feature for node \(i\), denoted by \(f_i'\), is computed as:

\begin{align}
f_i' = \sum_{j \in \mathcal{N}(i)} \alpha_{ij} Wf_j.
\end{align}

By dynamically adjusting each neighbor’s influence based on the computed attention weights, this approach allows the network to selectively emphasize informative connections while suppressing less relevant ones. This results in a more accurate representation of the event-based data, ultimately benefiting tasks like object recognition by enhancing robustness against noise.

\section{Performance Evaluation}

\vspace{2mm}
\begin{table*}[!t]
\centering
\captionsetup{justification=centering, singlelinecheck=false}
\renewcommand{\arraystretch}{1.3}
\caption{Comparison of object recognition accuracy across four datasets using various methods.}
\label{tab:comparison_results}
\begin{tabular}{@{}p{3cm}p{2.5cm}p{2cm}p{2cm}p{2cm}p{2cm}@{}}
\toprule
\textbf{Method} & \textbf{Representation} & \textbf{N-Caltech101} & \textbf{CIFAR10-DVS} & \textbf{MNIST-DVS} & \textbf{N-CARS} \\
\midrule
H-First & Spike & 5.4 & 7.7 & 59.5 & 56.1 \\
Gabor-SNN & Spike & 19.6 & 24.5 & 82.4 & 78.9 \\
HOTS & TimeSurface & 21.0 & 27.1 & 80.3 & 62.4 \\
HATS & TimeSurface & 64.2 & 52.4 & 98.4 & 90.2 \\
DART & TimeSurface & 66.4 & 65.8 & 98.5 & - \\
YOLE & VoxelGrid & 70.2 & - & 96.1 & 92.7 \\
AsyncNet & VoxelGrid & 74.5 & 66.3 & \textbf{99.4} & 94.4 \\
NVS-B & Graph & 67.0 & 60.2 & 98.6 & 91.5 \\
NVS-S & Graph & 67.0 & 60.2 & 98.6 & 91.5 \\
EvS-B & Graph & 76.1 & 68.0 & 99.1 & 93.1 \\
EvS-S & Graph & 76.1 & 68.0 & 99.1 & 93.1 \\
\midrule
\textbf{AW-GATCN (Ours)} & Graph & \textbf{83.77} & \textbf{76.79} & 99.3 & \textbf{96.89} \\
\bottomrule
\end{tabular}
\end{table*}
\vspace{-2mm}

\subsection{Experimental Setup}

We evaluate AW-GATCN on four event-based datasets: N-Caltech101, CIFAR10-DVS, MNIST-DVS, and N-CARS. Event data are segmented using adaptive windows based on normalized density and further voxelized via a square root law to balance spatial-temporal granularity. Graphs are constructed within each voxel with edge weights incorporating Euclidean distance, velocity, angular difference, and polarity consistency. Noise is filtered by maximizing the variance of the normalized degree distribution. AW-GATCN employs graph attention for feature aggregation, with fixed weight parameters $\alpha = 0.7$, $\beta = 0.1$, $\gamma = 0.1$, and $\delta = 0.1$. All models are trained using PyTorch with 400 epochs under 5-fold cross-validation. Accuracy is used as the evaluation metric.

\subsection{Comparison with Graph-based Methods for Object Recognition}

With optimal weight parameters (\(\alpha=0.7, \beta=0.1, \gamma=0.1, \delta=0.1\)), we evaluated our AW-GATCN model against state-of-the-art methods on four event-based object recognition benchmarks: N-Caltech101{}, CIFAR10-DVS{}, MNIST-DVS{}, and N-CARS{}. N-Caltech101, CIFAR10-DVS, and MNIST-DVS are derived from frame-based datasets by displaying moving images on a monitor and recording events with a fixed camera or monitor. N-Caltech101 matches the original Caltech101 in structure, with 8,246 samples across 101 classes. CIFAR10-DVS contains a sixth of the original CIFAR10 dataset, totaling 60,000 samples (6,000 per class). MNIST-DVS uses 10,000 symbols from MNIST, displayed at three scales, for a total of 30,000 samples. In contrast, N-CARS is captured directly with an event camera in real-world scenes, containing 12,336 car and 11,693 non-car samples.

As shown in Table~\ref{tab:comparison_results}, AW-GATCN achieved top-tier accuracy across all datasets, significantly outperforming existing approaches. Specifically, our model attained recognition accuracies of 83.77\%, 76.79\%, 99.3\%, and 96.89\% on N-Caltech101, CIFAR10-DVS, MNIST-DVS, and N-CARS, respectively. On challenging datasets such as N-Caltech101 and CIFAR10-DVS, AW-GATCN outperformed previous methods by a substantial margin, underscoring its robustness and effectiveness in complex, asynchronous event-based environments, where noise and heterogeneous data often pose significant challenges.

The results validate the effectiveness of the selected weight parameters from Experiment 1, which balanced edge factors to optimize both feature representation and noise reduction. The adaptive weighting approach allows AW-GATCN to dynamically adjust to diverse data characteristics, enabling it to perform well across different data domains.

In summary, these findings demonstrate the efficacy of AW-GATCN asynchronous event processing, establishing it as a robust and accurate model for event-based object recognition. By achieving high accuracy and resilience to noise across diverse settings, AW-GATCN shows significant potential to advance the state-of-the-art in event-driven applications.
\begin{figure*}[t]
    \centering

    \includegraphics[width=0.16\textwidth]{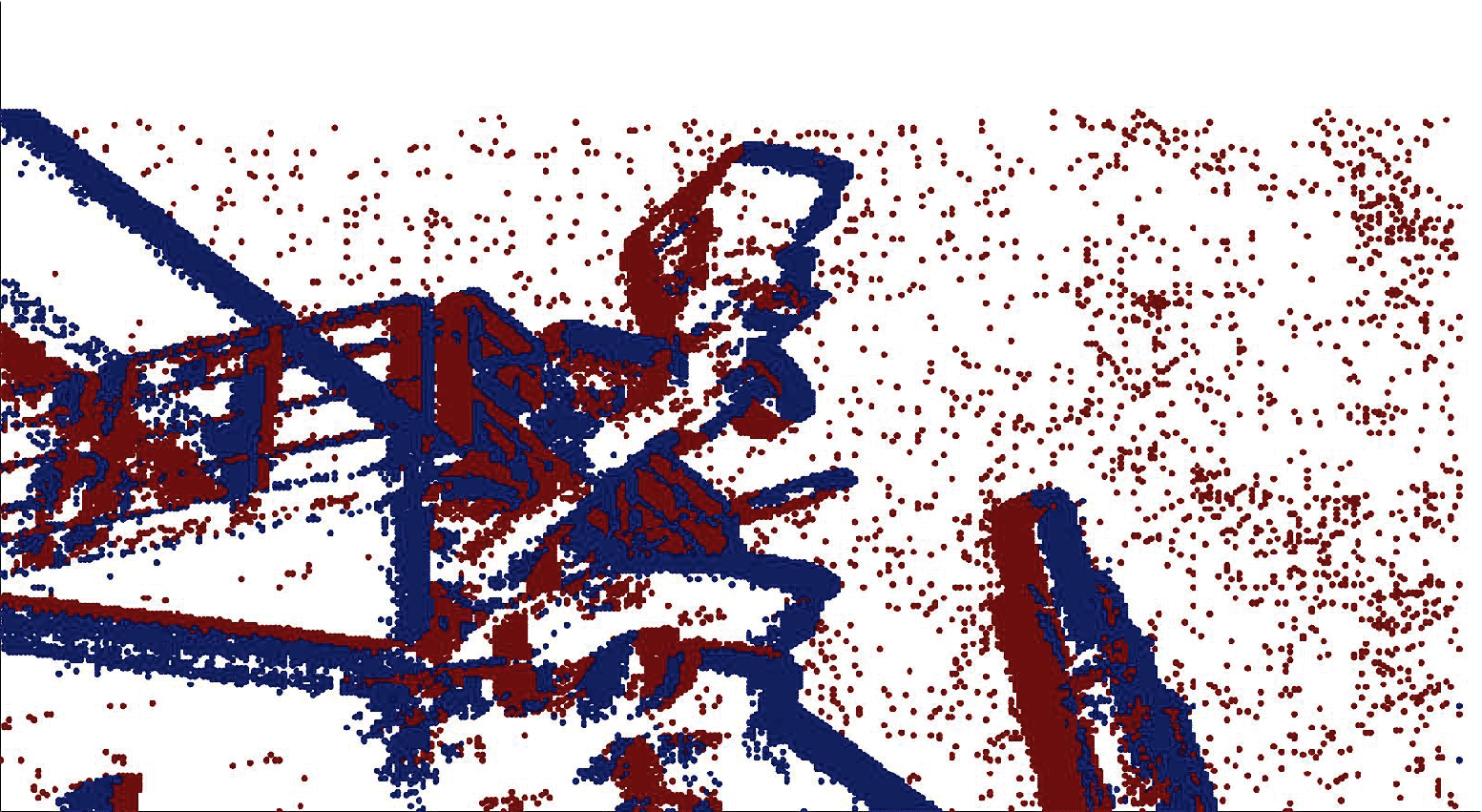}
    \hspace{0.1em}
    \includegraphics[width=0.16\textwidth]{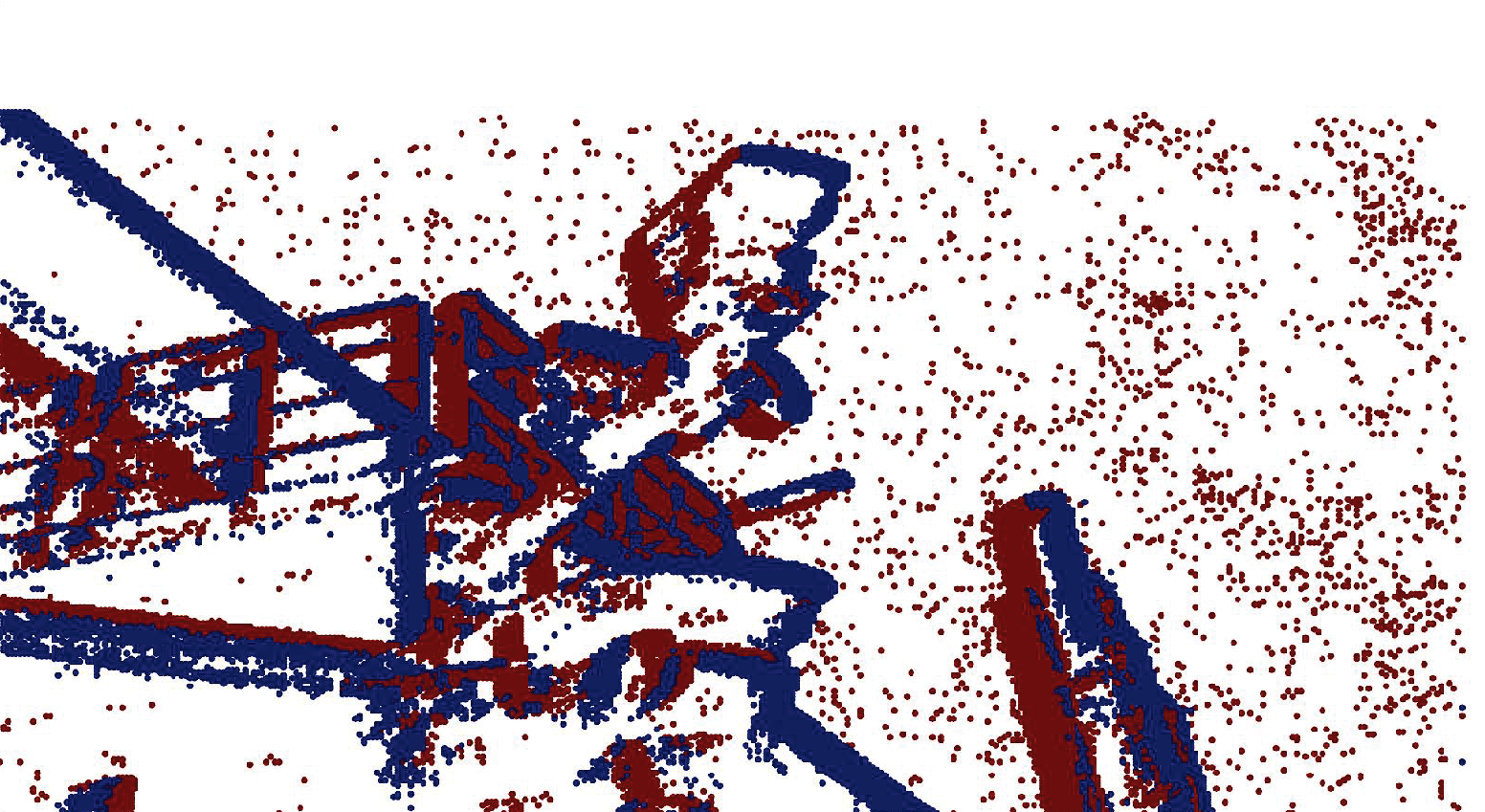}
    \hspace{0.1em}
    \includegraphics[width=0.16\textwidth]{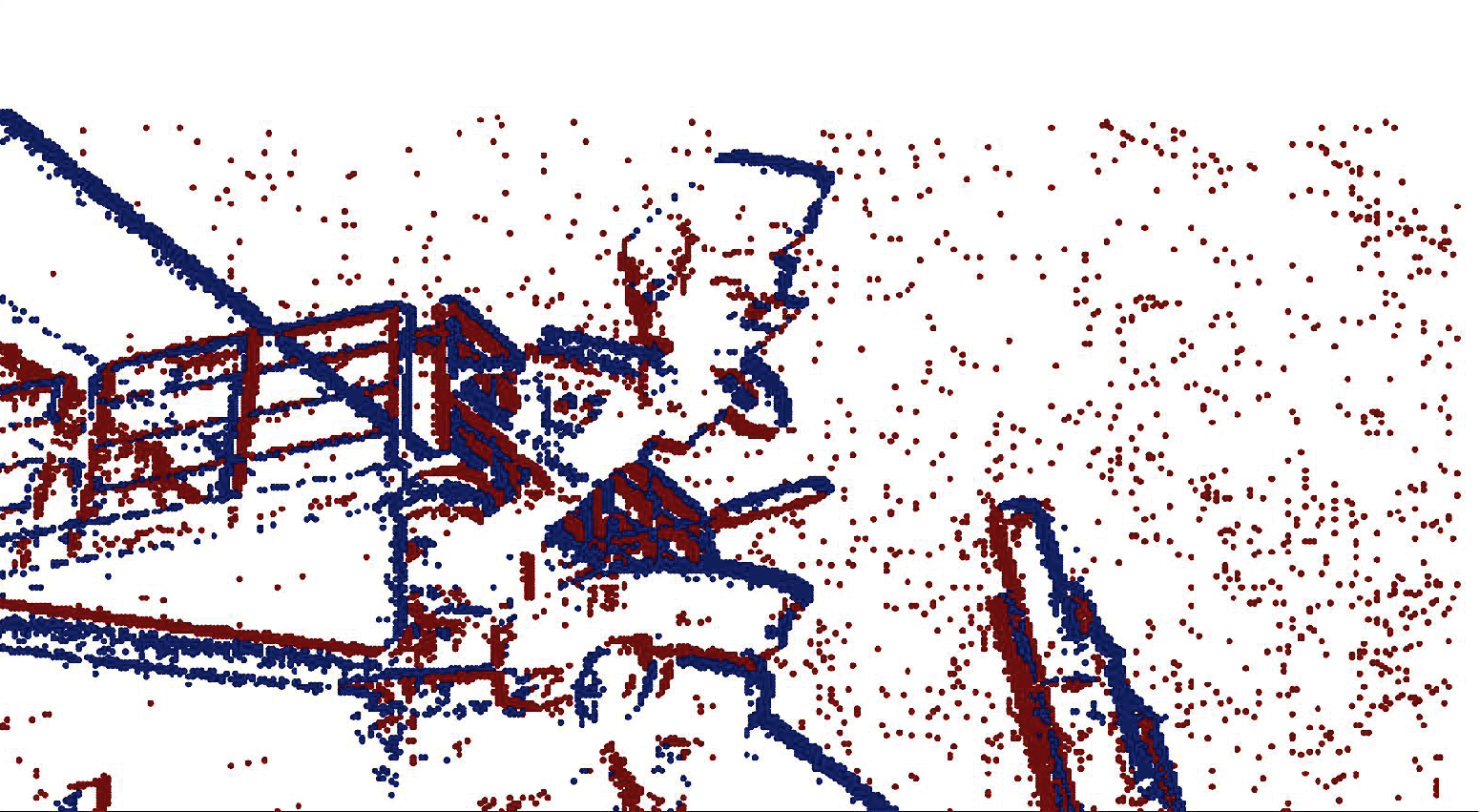}
    \hspace{0.1em}
    \includegraphics[width=0.16\textwidth]{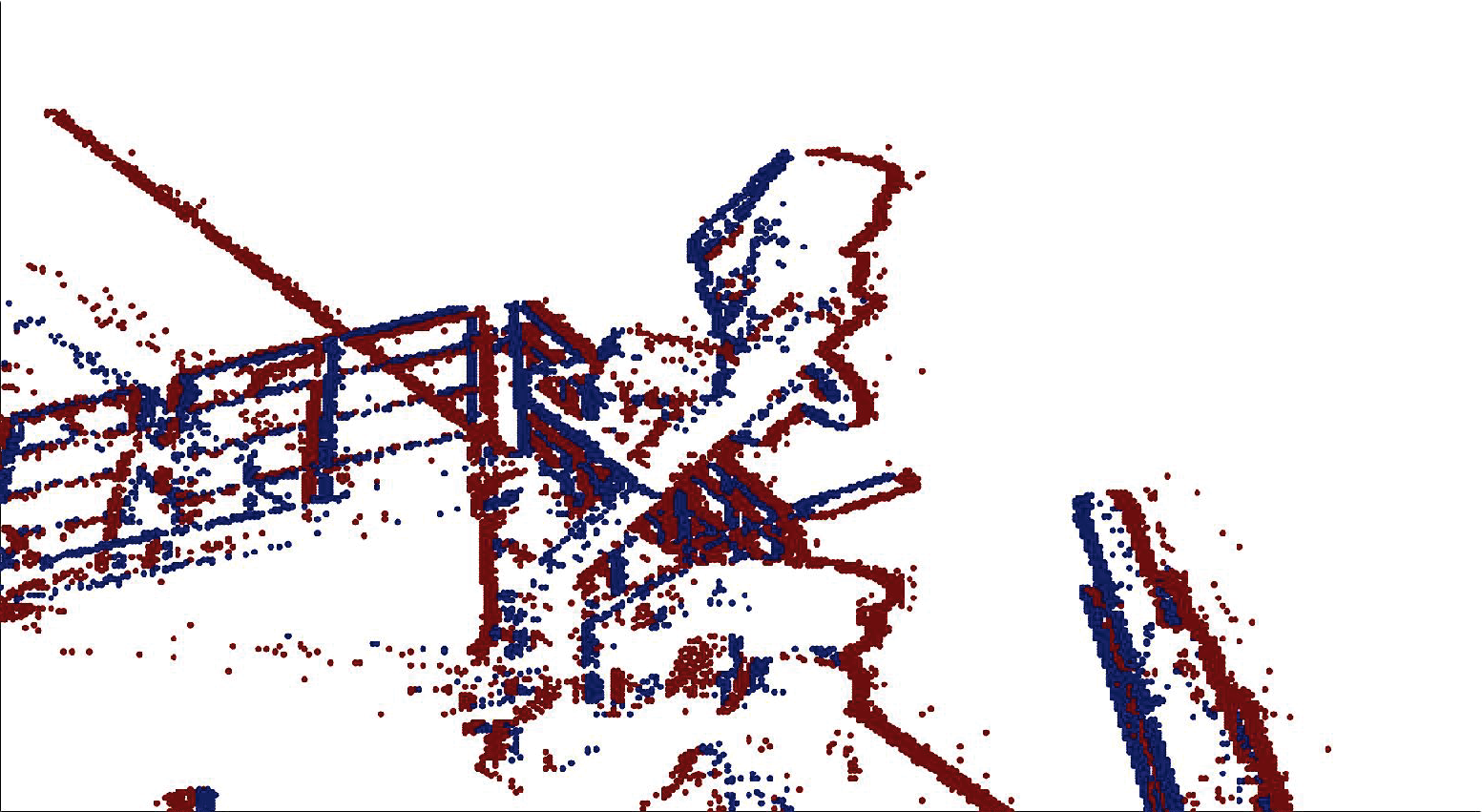}
    \hspace{0.1em}
    \includegraphics[width=0.16\textwidth]{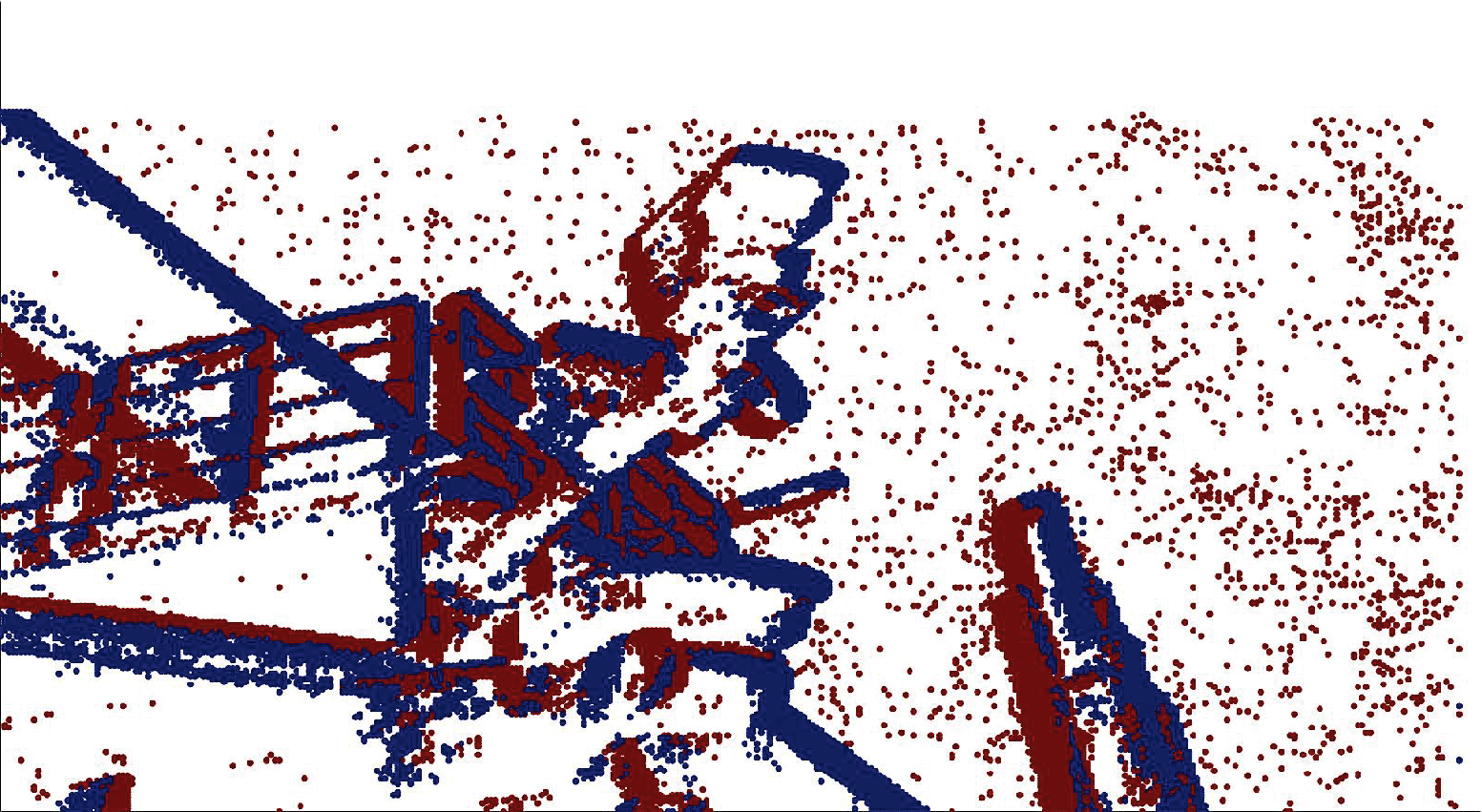}

    \vspace{0.3em} 
    \begin{tabular}{p{0.15\textwidth}p{0.15\textwidth}p{0.15\textwidth}p{0.15\textwidth}p{0.15\textwidth}}
        \centering (a) Original Data & \centering (b) Comb 1 & \centering (c) Comb 2 & \centering (d) Comb 3 & \centering (e) Comb 4 \\
    \end{tabular}

    \caption{Compared to Original Data, Comb 3 effectively reduces the number of event points while retaining the primary structure of the recognized object.}
    \label{fig:experiment_results}
\end{figure*}

\subsection{Ablation Study} 
\subsubsection{Determination of Weight Parameters}\label{sec:Determination of Weight Parameters}

We conducted experiments to optimize the weight parameters for multi-factor edge weighting across three datasets: N-END, D-END~\cite{b34}, and N-CARS, aiming to identify the optimal parameter configuration to maximize classification accuracy. The Event Noisy Dataset (END) consists of two parts: D-END (Daytime) and N-END (Night), providing diverse conditions to evaluate performance across different lighting environments.

The four evaluated weight configurations are as follows:
\begin{itemize}
    \item Comb 1: $\alpha=1, \beta=0, \gamma=0, \delta=0$
    \item Comb 2: $\alpha=0.8, \beta=0.1, \gamma=0.05, \delta=0.05$
    \item Comb 3: $\alpha=0.7, \beta=0.1, \gamma=0.1, \delta=0.1$
    \item Comb 4: $\alpha=0.6, \beta=0.2, \gamma=0.1, \delta=0.1$
\end{itemize}

These configurations reflect our approach, which leverages multiple factors for noise reduction rather than relying solely on Euclidean distance. Euclidean distance retains a higher weight as the primary criterion for noise determination, while additional factors provide supplementary information. Comb 2 assigns the highest auxiliary weight to velocity vector difference as it is the most relevant secondary factor. Comb 1, using only Euclidean distance, serves as a baseline.

Each configuration was tested using five-fold cross-validation, with the dataset split into five folds. Each fold was used as a test set once, while the remaining four folds formed the training set. The mean accuracy across the five runs was recorded for each configuration.

\begin{table}[h]
\centering
\caption{Recognition accuracy (\%) for various weight parameter combinations across different datasets (D-END, N-CARS, and N-END).}
\label{tab:weight_results}
\begin{tabular}{lcccc}
\toprule
\textbf{Dataset} & \textbf{Comb 1} & \textbf{Comb 2} & \textbf{Comb 3} & \textbf{Comb 4} \\
\midrule
D-END  & 89.15 & 91.73 & 93.65 & 90.81 \\
N-CARS & 90.63 & 94.64 & 96.89 & 91.57 \\
N-END  & 79.52 & 81.71 & 80.40 & 80.97 \\
\bottomrule
\end{tabular}
\end{table}

As shown in Table~\ref{tab:weight_results}, the configuration in Comb 3 achieved the highest accuracy on the D-END and N-CARS datasets, with scores of 93.65\% and 96.89\%, respectively. This balanced configuration, where Euclidean distance serves as the primary factor and auxiliary factors assist with noise reduction, proved particularly effective for complex datasets. For the N-END dataset, Comb 2 achieved slightly higher accuracy at 81.71\%, suggesting that emphasizing Euclidean distance is beneficial for datasets with similar characteristics.

As shown in Figure~\ref{fig:experiment_results}, Comb 3 effectively reduces the number of event points compared to other configurations, achieving an optimal balance between noise reduction and structural preservation. This configuration retains the essential characteristics of the object, allowing the network to capture critical spatial and temporal patterns necessary for accurate recognition. By selectively filtering out redundant or irrelevant data points, Comb 3 creates a more concise and refined representation of events, enhancing recognition performance without compromising the object’s primary structure. This balance between noise reduction and structural integrity accounts for Comb 3’s superior accuracy across most datasets.

These results indicate that while Comb 3’s balanced multi-factor approach generally provides optimal performance, adjusting the weight distribution, as in Comb 2, may yield improvements for specific datasets. Overall, a balanced configuration combining Euclidean distance with secondary factors is robust across diverse data conditions.

\subsubsection{Verification of Denoising Effectiveness}\label{tab:weight_results2}

To assess the impact of denoising, we conducted a comparative test of recognition accuracy using Comb 3, evaluating performance with and without denoising across the D-END, N-CARS, and N-END datasets. As shown in Table \ref{tab:weight_results2}, the denoising process significantly enhances recognition accuracy on all datasets. Specifically, denoising improved accuracy by 17.19\% on D-END and 19.57\% on N-CARS, demonstrating the robustness of AW-GATCN in complex background scenarios where noise can obscure essential features. 

For N-END, which primarily consists of data captured at night, the low-light conditions make the structure of objects less distinct, creating a complex noise environment. Denoising led to a 12.2\% improvement in accuracy, which, although smaller compared to more challenging datasets, underscores the generalizability of the denoising approach across various data complexities. By reducing noise interference, our model facilitates more accurate feature extraction, allowing the attention mechanism to prioritize meaningful connections over noise, which is essential for object recognition tasks in low-visibility scenarios.

These results validate the role of denoising in the AW-GATCN model, showing that the adaptive noise reduction strategy not only improves robustness in complex environments but also contributes to higher recognition accuracy across diverse data conditions. This ablation study underscores the effectiveness of incorporating denoising within a graph-based neural network framework, emphasizing its impact on model performance.

\begin{table}[h]
\centering
\caption{Recognition accuracy (\%) with and without denoising, using the Comb 3 weight configuration on D-END, N-CARS, and N-END datasets.}
\label{tab:weight_results2}
\begin{tabular}{lccc}
\toprule
\textbf{Comb Group} & \textbf{D-END} & \textbf{N-CARS} & \textbf{N-END} \\
\midrule
With Denoising & 93.65 & 96.89 & 80.40 \\
Without Denoising & 76.46 & 77.32 & 68.2 \\
Improvement & 17.19 & 19.57 & 12.2 \\
\bottomrule
\end{tabular}
\end{table}

\section{Conclusions}

We introduced AW-GATCN, an Adaptive Weighted Graph Attention Convolutional Network tailored for event-based data processing, excelling in denoising and object recognition. By integrating adaptive event point segmentation, multi-factor edge weighting, and an adaptive graph formulation-based noise reduction approach, AW-GATCN achieves superior accuracy and robustness, especially on noisy, heterogeneous event camera data. Experimental results show that AW-GATCN outperforms state-of-the-art methods with significant accuracy gains on challenging datasets. The optimized weight parameters and attention mechanism effectively prioritize essential connections, capturing spatiotemporal relationships that enhance noise resilience, feature aggregation, and recognition performance.

\end{document}